\documentclass[letterpaper,10pt,conference]{ieeeconf}

\IEEEoverridecommandlockouts %
\PassOptionsToPackage{table}{xcolor} %
\usepackage{amsmath} %
\usepackage{amssymb} %
\usepackage{amsfonts}
\usepackage[sort,compress]{cite}
\usepackage[usenames,dvipsnames]{xcolor}
\usepackage{graphicx}
\usepackage{tabularx}
\usepackage{multirow}
\usepackage{mathtools}
\usepackage{esvect} %
\usepackage[]{siunitx}
\usepackage{caption}
\usepackage[T1]{fontenc} %
\usepackage{mathtools, cuted} %
\usepackage{bm}
\usepackage{xspace}
\usepackage{arydshln}

\usepackage{etoolbox}
\usepackage[normalem]{ulem}
\usepackage{hyperref}
\robustify\bfseries
\robustify\uline

\newcolumntype{H}{>{\setbox0=\hbox\bgroup}c<{\egroup}@{}}

\usepackage{enumerate}
\usepackage{balance} %
\usepackage{booktabs} %

\newcommand{\NoRed}{\textbf{\textcolor{red}{No}}}
\newcommand{\YesGreen}{\textbf{\textcolor{ForestGreen}{Yes}}}

\newcommand{\OnlyPosRed}{\textbf{\textcolor{red}{Only Position}}}
\newcommand{\PosAndYawGreen}{\textbf{\textcolor{ForestGreen}{Position}} \& \textbf{\textcolor{ForestGreen}{Yaw}}}
\newcommand{\OptRed}{\textbf{Optimization-based} (\textbf{\textcolor{red}{slow}} \& \textbf{\textcolor{red}{not scalable}})}
\newcommand{\ILGreen}{\makecell{\textbf{IL-based} \\ (\textbf{\textcolor{ForestGreen}{faster}} \& \textbf{\textcolor{ForestGreen}{scalable}})}}

\usepackage{makecell}

\usepackage{amsmath,calc}

\usepackage{tikz} %
\usetikzlibrary{shapes,snakes} %
\usetikzlibrary{tikzmark} %
\usetikzlibrary{calc, arrows.meta, positioning} %
\usetikzlibrary{shapes.multipart} %
\usepackage{subcaption} %

\usepackage{footnote}
\makesavenoteenv{tabular}
\makesavenoteenv{table}

\ifdefined\qtyproduct
\else
  \ifdefined\NewCommandCopy
    \NewCommandCopy\qtyproduct\SI
  \else
    \NewDocumentCommand\qtyproduct{O{}mm}{\SI[#1]{#2}{#3}}
  \fi
\fi

\DeclareCaptionLabelSeparator{periodspace}{.\quad}
\newcommand{\trajANew}{traj\textsubscript{A\textsubscript{new}}}
\newcommand{\trajA}{traj\textsubscript{A}}

\definecolor{opt_color}{RGB}{203,255,182}
\definecolor{check_color}{RGB}{195,218,255}
\definecolor{recheck_color}{RGB}{255,176,176}
\definecolor{delaycheck_color}{RGB}{255,228,181}

\definecolor{agent1_color}{RGB}{234,153,153}
\definecolor{agent2_color}{RGB}{151,104,175}
\definecolor{agent3_color}{RGB}{249,203,156}
\definecolor{agent4_color}{RGB}{194,115,156}
\definecolor{agent5_color}{RGB}{159,197,232}
\definecolor{agent6_color}{RGB}{117,186,117}
\definecolor{obstacle_color}{RGB}{160,117,109}

\usepackage{mdframed}

\title{\LARGE \bf PRIMER: Perception-Aware Robust Learning-based \\%
Multiagent Trajectory Planner}

\author{Kota Kondo$^{1}$, Claudius T.\ Tewari$^{1}$, Andrea Tagliabue$^{1}$, Jesus Tordesillas$^{2}$, \\ Parker C.\ Lusk$^{1}$, Mason B. Peterson$^{1}$, Jonathan P.\ How$^{1}$%
	\thanks{$^{1}$The authors are with the Department of Aeronautics and Astronautics, Massachusetts Institute of Technology.
	    {\texttt{\{kkondo, cttewari, atagliab, plusk, kkondo, jhow\}@mit.edu.}}}
    \thanks{$^{2}$The author is with the Institute for Research in Technology, ICAI School of Engineering, Comillas Pontifical University (Spain). \tt{jtordesillas@comillas.edu}}
    \thanks{This work is supported by Boeing Research \& Technology and AFOSR MURI FA9550-19-1-0386}
}%

\begin{document}

\maketitle

\begin{abstract} 
In decentralized multiagent trajectory planners, agents need to communicate and exchange their positions to generate collision-free trajectories.
However, due to localization errors/uncertainties, trajectory deconfliction can fail even if trajectories are perfectly shared between agents.
To address this issue, we first present PARM and PARM*, perception-aware, decentralized, asynchronous multiagent trajectory planners that enable a team of agents to navigate uncertain environments while deconflicting trajectories and avoiding obstacles using perception information. PARM* differs from PARM as it is less conservative, using more computation to find closer-to-optimal solutions. While these methods achieve state-of-the-art performance, they suffer from high computational costs as they need to solve large optimization problems onboard, making it difficult for agents to replan at high rates. 
To overcome this challenge, we present our second key contribution, PRIMER, a learning-based planner trained with imitation learning (IL) using PARM* as the expert demonstrator. 
PRIMER leverages the low computational requirements at deployment of neural networks and achieves a computation speed up to $5614$ times faster than optimization-based approaches. 

\end{abstract}

\section*{Supplementary Material}
\noindent\textbf{Video}: \href{https://youtu.be/OmUxFmTrJOg}{https://youtu.be/OmUxFmTrJOg} \\

\section{Introduction}\label{sec:introduction}

In recent years, multiagent UAV trajectory planning has been extensively studied~\cite{ryou_cooperative_2022, peng2022obstacle, gao2022meeting, tordesillas2020mader, kondo2023robust, kondo2023robust_ral, zhou2020ego-swarm, sebetghadam2022distributed, robinson2018efficient, park2020efficient, Hou2022EnhancedDA, firoozi2020distributed, toumieh2023decentralized, wang2022robust, batra2022decentralized, csenbacslar2023mrnav, toumieh2024high, zhou2022swarm, csenbacslar2023dream, kondo2024puma, tordesillas2023deep, van2017distributed}. In real-world deployments of multiagent trajectory planning methods, it is crucial to deal with challenges such as (1) detecting and avoiding collisions with \textbf{unknown obstacles}, (2)~handling \textbf{localization errors/uncertainties}, (3)~achieving \textbf{scalability} to a large number of agents, and~(4) enabling \textbf{fast and efficient computation} for onboard replanning and quick adaptation to dynamic environments. However, finding effective solutions to these challenges remains an open question.

One approach to address challenges such as detecting and avoiding unknown obstacles, even in the presence of localization errors and uncertainties, is to equip each agent with a sensor, typically a camera, to perceive the surrounding environment. This allows agents to gather real-time information about their surroundings, enabling them to make informed decisions and take appropriate actions to avoid collisions and navigate through dynamic environments.
However, this sensor often has a limited field of view (FOV), making the orientation of the UAV crucial when planning trajectories through unknown space. Therefore, planners for flying with limited FOV sensors generally need to be perception-aware to ensure that as many obstacles or other UAVs as possible are kept within the FOV.

\begin{table}[!t]
    \renewcommand{\arraystretch}{1.4}
    \scriptsize
    \begin{centering}
    \caption{\centering State-of-the-art UAV Trajectory Planners}
    \label{tab:state_of_the_art_comparison}
    \resizebox{1.0\columnwidth}{!}{
    \begin{tabular}{>{\centering\arraybackslash}m{0.4\columnwidth} >{\centering\arraybackslash}m{0.25\columnwidth} >{\centering\arraybackslash}m{0.25\columnwidth} >{\centering\arraybackslash}m{0.25\columnwidth}}
    \toprule 
    \textbf{Method} & \textbf{Multiagent} & \textbf{Perception-aware} & \textbf{Learning-based} \tabularnewline
    \midrule
    \textbf{EGO-Swarm} \cite{zhou2020ego-swarm} & \cellcolor{LimeGreen!25} & \cellcolor{Red!10}  \tabularnewline
    \cline{0-0} 
    \textbf{DMPC} \cite{luis2020online} & \cellcolor{LimeGreen!25} & \cellcolor{Red!10} \tabularnewline
    \cline{0-0}
    \textbf{MADER} \cite{tordesillas2020mader} & \cellcolor{LimeGreen!25} & \cellcolor{Red!10} \tabularnewline
    \cline{0-0}
    \textbf{decMPC} \cite{toumieh2022decentralized} & \cellcolor{LimeGreen!25} & \cellcolor{Red!10} & \tabularnewline
    \cline{0-0}
    \textbf{Micro Swarm} \cite{zhou2022swarm} & \cellcolor{LimeGreen!25} \YesGreen{} & \cellcolor{Red!10} \NoRed{} & \textbf{No} \tabularnewline
    \cline{0-0}
    \textbf{RMADER} \cite{kondo2023robust, kondo2023robust_ral} & \cellcolor{LimeGreen!25} & \cellcolor{Red!10} \tabularnewline
    \cline{0-0}
    \textbf{MRNAV} \cite{csenbacslar2023mrnav} & \cellcolor{LimeGreen!25} & \cellcolor{Red!10} \tabularnewline
    \cline{0-0}
    \textbf{DREAM} \cite{csenbacslar2023dream} & \cellcolor{LimeGreen!25} & \cellcolor{Red!10} \tabularnewline
    \cline{0-0}
    \textbf{HDSM} \cite{toumieh2024high} & \cellcolor{LimeGreen!25} & \cellcolor{Red!10} \tabularnewline
    \hline
    \textbf{Raptor} \cite{zhou2021raptor} & \cellcolor{Red!10} & \cellcolor{LimeGreen!25} \tabularnewline
    \cline{0-0}
    \textbf{Time-opt} \cite{spasojevic2020perception} & \cellcolor{Red!10} & \cellcolor{LimeGreen!25} \tabularnewline
    \cline{0-0}
    \textbf{PANTHER} \cite{tordesillas2022panther} & \cellcolor{Red!10} \NoRed{} & \cellcolor{LimeGreen!25} \YesGreen{} & \textbf{No} \tabularnewline
    \cline{0-0}
    \textbf{PA-RHP} \cite{wu2022perception} & \cellcolor{Red!10} & \cellcolor{LimeGreen!25} \tabularnewline
    \cline{0-0}
    \textbf{PA-RHP} \cite{falanga2018pampc} & \cellcolor{Red!10} & \cellcolor{LimeGreen!25} \tabularnewline
    \hline
    \textbf{Deep-PANTHER} \cite{tordesillas2023deep} & \cellcolor{Red!10} & \cellcolor{LimeGreen!25} & \tabularnewline
    \cline{0-0}
    \textbf{CGD} \cite{kondo2024cgd} & \cellcolor{Red!10}  \NoRed{} & \cellcolor{LimeGreen!25} \YesGreen{} & \textbf{Yes} \tabularnewline    
    \cline{0-0}
    \textbf{LPA} \cite{song2023learning} & \cellcolor{Red!10}  & \cellcolor{LimeGreen!25} \tabularnewline    
    \hline
    \textbf{PUMA} \cite{kondo2024puma} & \cellcolor{LimeGreen!25}  & \cellcolor{LimeGreen!25} & \textbf{No} \tabularnewline
    \cline{0-0}
    \textbf{PARM} (proposed) & \cellcolor{LimeGreen!25} \YesGreen{} & \cellcolor{LimeGreen!25} \YesGreen{} & \textbf{No} \tabularnewline
    \cline{0-0}
    \textbf{PRIMER}\ (proposed) & \cellcolor{LimeGreen!25} & \cellcolor{LimeGreen!25}  & \textbf{Yes} \tabularnewline
    \bottomrule
    \end{tabular}}
    \par\end{centering}
    \vspace{-20pt}
\end{table}

When scaling multiagent trajectory planners, it is important to note that, with centralized planners, each agent needs to listen to a single entity that plans all the trajectories~\cite{robinson2018efficient, park2020efficient}. While this approach simplifies planning, the central entity may act as a single point of failure, and the replanning abilities of the agent depend on their ability to communicate with the central entity. Decentralized planners greatly mitigate these issues, as each agent plans its own trajectory \cite{zhou2020ego-swarm, tordesillas2020mader, sebetghadam2022distributed, kondo2023robust, toumieh2023decentralized, batra2022decentralized}. Decentralized planners are therefore generally considered to be inherently more scalable and robust to failures.

Similarly, synchronous planners such as~\cite{van2017distributed, sebetghadam2022distributed, firoozi2020distributed} require all agents to wait at a synchronization barrier until planning can be globally triggered, whereas asynchronous planning enables each agent to independently trigger the planning step without considering the planning status of other agents. Asynchronous methods are typically more scalable compared to synchronous methods~\cite{zhou2020ego-swarm, kondo2023robust, tordesillas2020mader}.

Many optimization-based approaches~\cite{tordesillas2020mader, kondo2023robust, toumieh2023decentralized, zhou2020ego-swarm, kondo2024puma} have been proposed for multiagent trajectory generation. However, these methods often require significant computational resources, making them challenging to deploy in dynamic environments where fast, on-the-fly replanning is essential. In particular, when optimizing perception-aware trajectories while considering multiagent interactions, computation times increase significantly, as demonstrated in Section~\ref{sec:simultation-results} and \cite{kondo2024puma}, which hinders real-time deployment. To address this issue, researchers have explored imitation learning (IL)-based approaches~\cite{tordesillas2023deep, Tagliabue2021DemonstrationEfficientGP, Park2020VisionBasedOA}, which offer the advantage of faster replanning while still producing near-optimal trajectory generation.

To tackle the challenges of (1)~\textbf{unknown objects detection and collision avoidance}, (2)~\textbf{localization errors/uncertainties}, (3)~\textbf{scalability}, and (4)~\textbf{fast and efficient computation}, we propose PRIMER, an IL-based decentralized, asynchronous, perception-aware multiagent trajectory planner. Table~\ref{tab:state_of_the_art_comparison} provides a comparison of PRIMER with state-of-the-art approaches, and our contributions include:
\begin{enumerate}
    \item PARM/PARM* -- decentralized, asynchronous, perception-aware, multiagent trajectory planner.
    \item PRIMER -- IL-based decentralized, asynchronous, perception-aware, multiagent approach for translational and yaw trajectory generation with the use of Long Short-Term Memory (LSTM) neural network.
    \item Extensive simulation benchmarking, we compared the performance of PARM, PARM*, and PRIMER.
\end{enumerate}

\section{Perception-aware Multiagent Trajectory Generation}\label{sec:trajectory-generation}

This section outlines our perception-aware multiagent planning approach, encompassing both optimization-based and IL-based methods.
First, we introduce PARM and PARM*, which are optimization-based perception-aware multiagent trajectory planners. These planners utilize the Robust MADER trajectory deconfliction framework~\cite{kondo2023robust_ral} to ensure safe and collision-free trajectories. By considering both position and yaw, PARM and PARM* are capable of tracking multiple obstacles while generating trajectories.
Next, we present PRIMER, an IL-based planner that is trained using PARM* as an expert. PRIMER offers fast computation by leveraging the benefits of IL. By incorporating the expertise of PARM*, PRIMER is able to generate effective trajectories while significantly reducing computational requirements.
As Table~\ref{tab:state_of_the_art_pa_comparison} shows PRIMER is the first learning-based perception-aware multiagent trajectory planner designed to track multiple obstacles while accounting for both position and yaw.

\subsection{PARM / PARM* \textemdash Optimization-based Decentralized, Asynchronous Perception-Aware Multiagent Planning}

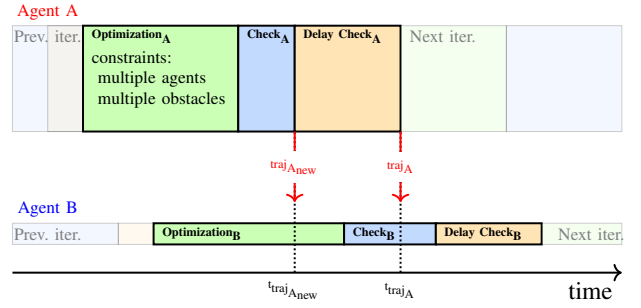
\begin{figure}[t]
    \centering
    \resizebox{\columnwidth}{!}{
    \begin{tikzpicture}
[
greenbox/.style={shape=rectangle, fill=opt_color, draw=black},
bluebox/.style={shape=rectangle, fill=check_color, draw=black},
 yellowbox/.style={shape=rectangle, fill=delaycheck_color, draw=black},
]

\newcommand\Ay{3.5}
\newcommand\Axo{1}
\newcommand\Axc{3.2}
\newcommand\Axr{4}
\newcommand\Axe{5.5}

\newcommand\By{0.7}
\newcommand\Bxo{2.0}
\newcommand\Bxc{4.7}
\newcommand\Bxr{6.0}
\newcommand\Bxe{7.5}

    \node[text=red] at (0.5,\Ay+0.2) {\scriptsize Agent A};
    \filldraw[fill=check_color, draw=black, opacity=0.2] (0,\Ay) rectangle (1.5,\Ay-1.5);
    \filldraw[fill=delaycheck_color, draw=black, opacity=0.2] (0.5,\Ay) rectangle (\Axo,\Ay-1.5);
    \filldraw[thick, fill=opt_color, draw=black] (\Axo,\Ay) rectangle (\Axc,\Ay-1.5);
    \filldraw[thick, fill=check_color, draw=black] (\Axc, \Ay) rectangle (\Axr, \Ay-1.5);
    \filldraw[thick, fill=delaycheck_color, draw=black] (\Axr, \Ay) rectangle (\Axe, \Ay-1.5);
    \filldraw[fill=opt_color, draw=black, opacity=0.2] (\Axe, \Ay) rectangle (\Axe+1.5, \Ay-1.5);
    \filldraw[fill=check_color, draw=black, opacity=0.2] (\Axe+1.5, \Ay) rectangle (\columnwidth, \Ay-1.5);
    \node[text=blue] at (0.5,\By+0.2) {\scriptsize Agent B};
    \filldraw[fill=check_color, draw=black, opacity=0.2] (0,\By) rectangle (\Bxo-0.5,\By-0.3);
    \filldraw[fill=delaycheck_color, draw=black, opacity=0.2] (\Bxo-0.5,\By) rectangle (\Bxo,\By-0.3);
    \filldraw[thick, fill=opt_color, draw=black] (\Bxo,\By) rectangle (\Bxc,\By-0.3);
    \filldraw[thick, fill=check_color, draw=black] (\Bxc, \By) rectangle (\Bxr, \By-0.3);
    \filldraw[thick, fill=delaycheck_color, draw=black] (\Bxr, \By) rectangle (\Bxe, \By-0.3);
    \filldraw[fill=opt_color, draw=black, opacity=0.2] (\Bxe, \By) rectangle (\columnwidth, \By-0.3);

\draw[thick, densely dotted] (\Axr,0) -- (\Axr,\Ay-1.5) node[] at (\Axr, -0.25) {\tiny t\textsubscript{traj\textsubscript{A\textsubscript{new}}}};
\draw[thick, densely dotted] (\Axe,0) -- (\Axe,\Ay-1.5) node[] at (\Axe, -0.25) {\tiny t\textsubscript{traj\textsubscript{A}}};
    
\draw[thick,->] (0,0) -- (\columnwidth,0) node[anchor=north east] {time};

\draw[thick, ->, draw=red] (\Axr,\Ay-1.5) -- (\Axr,\Ay-2.5) node[midway,fill=white, text=red] {\tiny \trajANew{}};
\draw[thick, ->, draw=red] (\Axe,\Ay-1.5) -- (\Axe,\Ay-2.5) node[midway,fill=white, text=red] {\tiny \trajA{}};

\node[font=\bfseries,right] at (\Axo,\Ay-0.15) {\tiny Optimization\textsubscript{A}};
\node[right] at (\Axo,\Ay-0.45) {\scriptsize constraints:};
\node[right] at (\Axo,\Ay-0.75) {\scriptsize \ multiple agents};
\node[right] at (\Axo,\Ay-1.05) {\scriptsize \ multiple obstacles};
\node[font=\bfseries,right] at (\Axc,\Ay-0.15) {\tiny Check\textsubscript{A}};
\node[font=\bfseries,right] at (\Axr,\Ay-0.15) {\tiny Delay Check\textsubscript{A}};

\node[font=\bfseries,right] at (\Bxo,\By-0.15) {\tiny Optimization\textsubscript{B}};
\node[font=\bfseries,right] at (\Bxc,\By-0.15) {\tiny Check\textsubscript{B}};
\node[font=\bfseries,right] at (\Bxr,\By-0.15) {\tiny Delay Check\textsubscript{B}};

\node[color=gray] at (0.5,\Ay-0.15) {\scriptsize Prev. iter.};
\node[color=gray] at (\Axe+0.6,\Ay-0.15) {\scriptsize Next iter.};
\node[color=gray] at (0.5,\By-0.15) {\scriptsize Prev. iter.};
\node[color=gray] at (0.95\columnwidth,\By-0.15) {\scriptsize Next iter.};

\end{tikzpicture}
    } 
    \captionof{figure}{PARM 's trajectory optimization and deconfliction sequence: PARM uses an optimization-based approach to generate trajectories for each agent, followed by a conflict detection and resolution step based on the Robust MADER framework. Each agent first generates a new trajectory in the optimization step, and then checks if there are any conflicts with the trajectories received from other agents. If no conflicts are detected, the agent publishes its new trajectory and begins checking for potential collisions in a delay check step. This delay check step is a sequence of checks over a period of time. Finally, if no conflicts are detected during the delay check, the agent commits to the new trajectory and publishes it. However, if conflicts are detected, the agent reverts to the trajectory from the previous iteration and discards the new trajectory. More details on the RMADER approach can be found in Section II of~\cite{kondo2023robust_ral}.}
    \vspace{-20pt}
\label{fig:parm_sequence}
\end{figure}

Our previous work, PANTHER~\cite{tordesillas2022panther}, proposed a perception-aware trajectory planner for a single agent in dynamic environments, generating trajectories to avoid obstacles while keeping them in the sensor FOV. In~\cite{tordesillas2023deep}, PANTHER* improved the original PANTHER with less conservatism by including separating planes and trajectory total time as optimization variables, but both were limited to tracking and avoiding only one obstacle at a time.
To address this limitation, we modify the optimization problem solved by PANTHER to enable the tracking and avoidance of multiple obstacles, resulting in PARM—a decentralized, asynchronous, perception-aware multi-agent trajectory planning system. PARM incorporates this modified optimization approach into the RMADER~\cite{kondo2023robust_ral} deconfliction framework, which solves a feasibility linear problem between its own committed trajectories and the shared trajectories from other agents (see Fig.~\ref{fig:parm_sequence} and \cite{tordesillas2020mader, kondo2023robust_ral} for details). To facilitate the tracking of multiple obstacles and agents, we modify the FOV term as described in Section IV of \cite{tordesillas2022panther} as follows.
\begin{equation}
    - \alpha_{FOV} \sum_{i}^{n} \left[ \int_{0}^{T} \{\text{inFOV}(\text{obstacle}_{i})\}^3 dt \right] \\
\end{equation}
where $\alpha_{FOV}$ is the weight for the FOV term, $n$ is the number of obstacles, $T$ is the total time of the trajectory, inFOV() returns a higher number when obstacle$_i$ is in FOV.

Additionally, following the method outlined in~\cite{tordesillas2023deep}, which minimizes trajectory time in the optimization formulation and consequently reduces trajectory costs, we introduce PARM*, a method that generates closer-to-optimal trajectories. However, this more complex formulation increases computation time, resulting in slower replanning rates.

Note that coupled trajectory generation has the advantage of enabling the optimizer to simultaneously consider both position and yaw trajectories, taking into account the impact of each on the other. On the other hand, the decoupled approach involves optimizing either the position or yaw trajectory first, and then optimizing the other trajectory based on the pre-determined trajectory. While this approach may reduce computation and complexity, it could also result in sub-optimal position and yaw trajectories overall, as the two trajectories are not jointly optimized. Therefore we designed PARM* in a way that they optimize both position and yaw in a coupled manner (See Table~\ref{tab:state_of_the_art_pa_comparison}). 

Note that we assume our system is differential flat~\cite{mellinger2011minimum}, and the flat outputs are position, yaw, and their derivatives.

\subsection{PRIMER \textemdash Imitation Learning-based PARM*}

In our previous work, Deep-PANTHER \cite{tordesillas2023deep}, we used IL to train a neural network that generates a desired position trajectory, while relying on closed-form solutions to obtain the direction where the onboard sensor should be looking at (e.g., yaw on a multirotor). This closed-form yaw solution generates yaw trajectories given position trajectories, reducing the output dimension of the learned policy. However, this approach is not scalable in multi-obstacle environments since the closed-form solution only generates yaw trajectories for a single given obstacle.
To address this limitation, we design PRIMER using a multi-layer perceptron (MLP) that generates both position and yaw trajectories. 
To generate both position and yaw trajectories, PRIMER has the size of the neural network to $4$ fully connected layers, each with $1024$ neurons, and is trained to imitate the optimal perception-aware trajectories generated by PARM*.

Additionally, we added a Long Short-Term Memory (LSTM)~\cite{HochSchm97} feature-extraction network to the MLP used in PRIMER, inspired by the ground-robot motion planning approach \cite{everett2018motion}. 
This allowed the neural network to accept various numbers of obstacles and agents as input, whereas traditional fully-connected feedforward neural networks can only handle a fixed number of obstacles. LSTM can take as many obstacles and agents as possible and generate a fixed length of the latent output, which we feed into the fully connected layers. 

It is also worth noting that IL-based approaches are more scalable in practice than optimization-based approaches. As the number of agents and obstacles in the environment increases, optimization-based approaches need to include more constraints in the optimization, leading to significant computational requirements. On the other hand,  IL-based approaches such as PRIMER are able to handle larger-scale environments with little to no additional computational overhead with the use of LSTM.

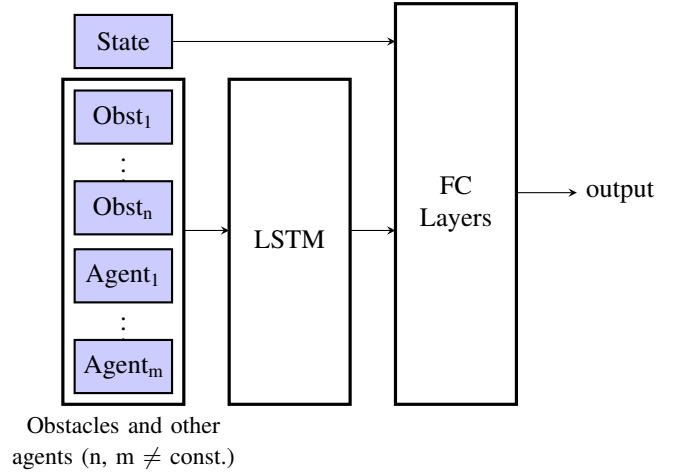
\begin{figure}[h]
    \centering
    \begin{tikzpicture}
    [block_o/.style = {rectangle, draw=black, thick, width=5em, height=10em},
    block_on/.style = {rectangle, draw=black, fill=white!80!blue, thick, text width=3em,align=center, minimum height=2em}]

    \draw (-2,3.5) node[block_on] (A) {State};

    \draw[black, very thick] (-2.8,-1.3) rectangle (-1.2,3);
    \draw (-2,-1.8) node[] {\makecell{\small Obstacles and other \\ \small agents (n, m $\neq$ const.)}};
    \draw (-2,2.5) node[block_on] (O1) {Obst\textsubscript{1}};
    \draw (-2,1.9) node[] {\vdots};
    \draw (-2,1.3) node[block_on] (On) {Obst\textsubscript{n}};
    \draw (-2,0.4) node[block_on] (A1) {Agent\textsubscript{1}};
    \draw (-2,-0.2) node[] {\vdots};
    \draw (-2,-0.8) node[block_on] (Am) {Agent\textsubscript{m}};

    \draw[black, very thick] (-0.6,-1.3) rectangle (1.0,3) node[midway, anchor=center] {\makecell{LSTM}};

    \draw[black, very thick] (1.6,-1.3) rectangle (3.2,4) node[midway, anchor=center] {\makecell{FC \\ Layers}};
    
    \draw[-stealth] (A) -- (1.6,3.5);
    \draw[-stealth] (-1.2, 1) -- (-0.6, 1);
    \draw[-stealth] (1.0, 1) -- (1.6, 1);
    \draw[-stealth] (3.2, 1.5) -- (4.0, 1.5) node[anchor=west] {output};

\end{tikzpicture}
\caption{\centering PRIMER Network Architectures}
\label{fig:primer-nn}
\vspace{-5pt}
\end{figure}

Fig.~\ref{fig:primer-nn} shows an architecture of PRIMER. We first feed the predicted trajectories of obstacles and received other agents' trajectories to the LSTM, which outputs a fixed size vector.
We then combine it with the agent's own state and feed this into the fully connected layers.

Table~\ref{tab:state_of_the_art_pa_comparison} shows the comparison of the state-of-the-art perception-aware trajectory planners. PARM/PARM* are the first perception-aware multiagent trajectory planner that generates position and yaw coupled trajectory while tracking multiple obstacles, and PRIMER is the extension of PARM, in the sense that it has much faster computation time, leveraging IL-based planner.

Similar to PARM/PARM*, PRIMER leverages Robust MADER's trajectory deconfliction approach, ensuring that all trajectories generated by PRIMER undergo deconfliction checks to guarantee collision-free trajectory generation between agents. Unlike traditional optimization-based methods, which ensure collision-free trajectories by incorporating trajectories from other agents into their constraints, NN-based planners do not inherently provide such guarantees. However, the overall architecture ensures that the final committed trajectories at the conclusion of the replanning steps are collision-free.

\begin{table}[h]
    \renewcommand{\arraystretch}{1}
    \scriptsize
    \begin{centering}
    \caption{\centering State-of-the-art Perception-aware Obstacle Tracking Trajectory Planners}
    \label{tab:state_of_the_art_pa_comparison}
    \resizebox{0.9\columnwidth}{!}{
    \begin{tabular}{>{\centering\arraybackslash}m{0.16\columnwidth} || >{\centering\arraybackslash}m{0.1\columnwidth} >{\centering\arraybackslash}m{0.06\columnwidth} >{\centering\arraybackslash}m{0.2\columnwidth} >{\centering\arraybackslash}m{0.26\columnwidth}} 
    \toprule 
    \textbf{Method} & \textbf{Tracking Multi-obstacles} & \makecell{\textbf{Multi-} \\ \textbf{agents}} & \textbf{Trajectory} & \textbf{Planning} \tabularnewline
    \midrule
    \cite{thomas2017autonomous} & \NoRed{} & \NoRed{} & \OnlyPosRed{}{} & \OptRed{} \tabularnewline
    \hline
    \cite{penin2017vision-based} & \NoRed{} & \NoRed{} & \PosAndYawGreen{} & \OptRed{} \tabularnewline
    \hline
    \textbf{PANTHER} / \textbf{PANTHER*} \cite{tordesillas2022panther, tordesillas2023deep} & \NoRed{} & \NoRed{} & \PosAndYawGreen{} & \OptRed{} \tabularnewline
    \hline
    \textbf{Deep-PANTHER} \cite{tordesillas2023deep} & \NoRed{} & \NoRed{} & \OnlyPosRed{} \footnote{Deep-PANTHER \cite{tordesillas2023deep} generates only position trajectory, and yaw trajectory is generated by closed-form solution based on the position trajectory.} & \ILGreen{} \tabularnewline
    \hline
    \textbf{PARM} / \textbf{PARM*} (proposed) & \YesGreen{} & \YesGreen{} & \PosAndYawGreen{} & \OptRed{} \tabularnewline
    \hline
    \textbf{PRIMER} (proposed) & \YesGreen{} & \YesGreen{} & \PosAndYawGreen{} & \ILGreen{} \tabularnewline
    \bottomrule
    \end{tabular}}
    \par\end{centering}
\vspace{-15pt}
\end{table}

\subsection{PRIMER Training Setup}
We used the student-expert IL learning framework, where PARM* acts as an expert that provides demonstrations, and PRIMER is the student, trained so that its neural network can reproduce the provided demonstrations. 
We trained the student in an environment that contained multiple dynamic obstacles flying a randomized trefoil-knot trajectory, as well as other PRIMER agents. The terminal goal for the student was also randomized. To collect the data and train the student, we utilized the Dataset-Aggregation algorithm (DAgger)~\cite{ross2011reduction}, and Adam~\cite{kingma2014adam} optimizer. Additionally, we introduced a weighted loss function between position and yaw. During the training process, we found that it was more difficult to train the yaw trajectory than the position trajectory, and thus we weighted the yaw loss function. In our training, we set the weight $\alpha$ to $70$. The total loss is defined as:
\begin{equation}
\mathcal{L}_\text{total} = \mathcal{L}_\text{pos} + \alpha \mathcal{L}_\text{yaw}
\end{equation}
where $\mathcal{L}_\text{total}$ is the total loss, $\mathcal{L}_\text{pos}$ is the loss for the position, $\alpha$ is the weight between the position and yaw trajectories, and $\mathcal{L}_\text{yaw}$ is the loss for the yaw trajectory.

\subsection{Obstacle Sharing}
\begin{figure}[h]
    \centering
    \includegraphics[width=\columnwidth, trim={0 6cm 0 0.5cm}, clip]{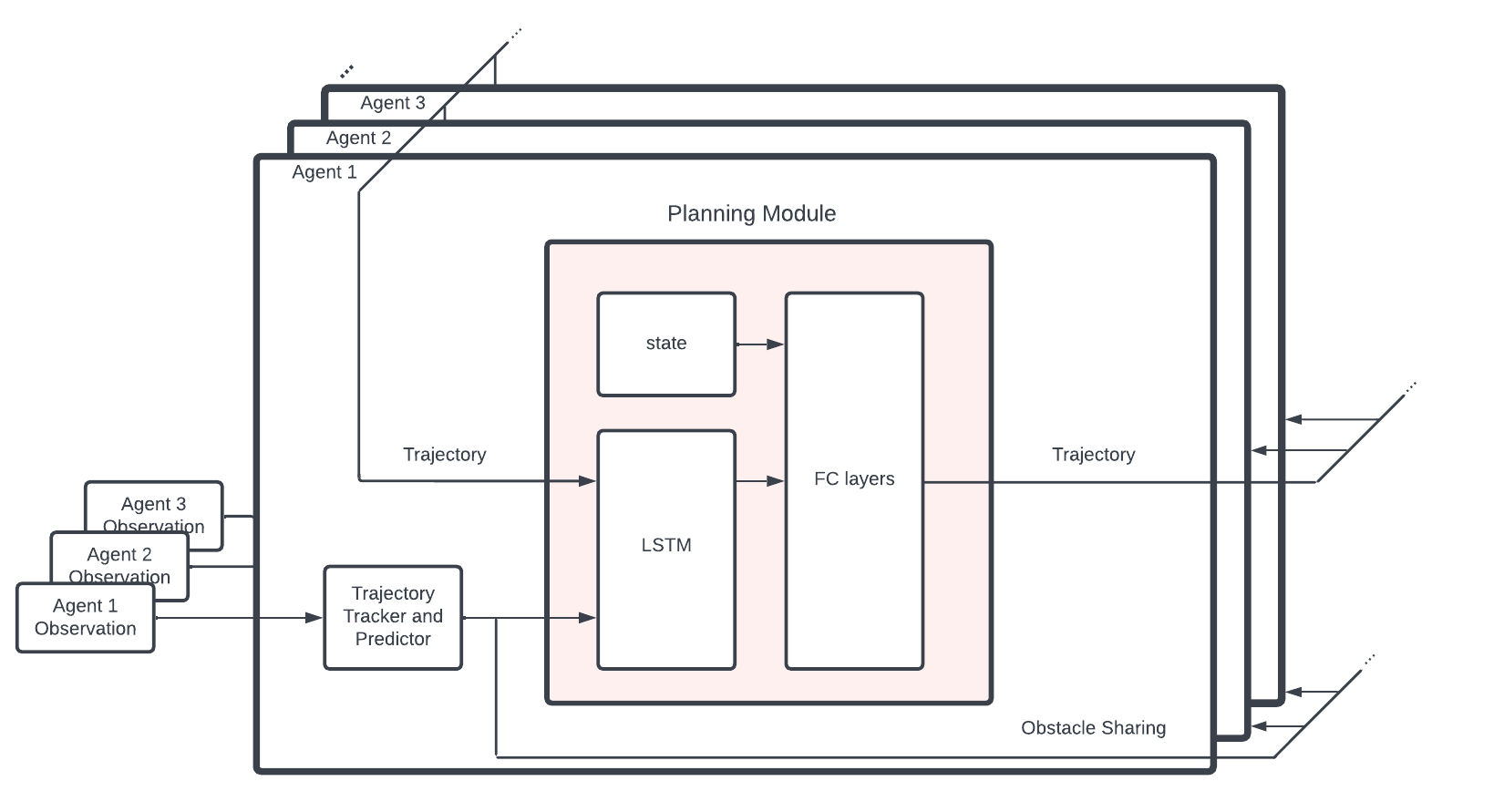}
\caption{\centering PRIMER Planning and Sharing Trajectory Architecture}
\label{fig:primer-planning-sharing-architecture}
\vspace{-10pt}
\end{figure}

As shown in Fig.~\ref{fig:primer-planning-sharing-architecture}, each agent detects and tracks obstacles and shares their predicted trajectories with other agents. This obstacle-sharing architecture allows the agents to have a better understanding of the surrounding environment as a team.

\section{Simulation Results}\label{sec:simultation-results}

\subsection{PRIMER vs. PARM* in single-agent, single-obstacle environment} 

Table~\ref{tab:sim-compare-primer-parm_star} compares the average performance of PRIMER and PARM* in a simulation environment with a single dynamic obstacle following a trefoil trajectory, while the agent flies diagonally to avoid obstacles. The comparison is based on the average trajectory cost, computation time, obstacle avoidance failure rate, and dynamic constraint failure rate. As shown in Table~\ref{tab:sim-compare-primer-parm_star}, PRIMER achieves a \textbf{5614-time} reduction in computation time, with only a \SI{6.7}{\%} increase in cost, while maintaining a \SI{100}{\%} success rate and \SI{0}{\%} dynamic constraint violations.

\begin{table}[h]
\renewcommand{\arraystretch}{1.4}
\caption{\centering PRIMER vs. PARM* in single-agent, single-obstacle environment}
\label{tab:sim-compare-primer-parm_star}
\begin{centering}
\resizebox{0.9\columnwidth}{!}{
\begin{tabular}{>{\centering\arraybackslash}m{0.14\columnwidth} || >{\centering\arraybackslash}m{0.1\columnwidth} >{\centering\arraybackslash}m{0.1\columnwidth} >{\centering\arraybackslash}m{0.08\columnwidth} >{\centering\arraybackslash}m{0.3\columnwidth}}
\toprule
 & \textbf{Avg. Cost} & \textbf{Compu. Time [ms]} & \textbf{Success Rate [\%]} & \textbf{Obst. Avoidance, Dyn. Constr. Failure Rate [\%]} \tabularnewline
\midrule
\textbf{PARM*} & 403 & \textbf{\textcolor{Red}{4495.8}} & \textbf{\textcolor{ForestGreen}{100}} & 0.0  \tabularnewline
\hline 
\textbf{PRIMER} & 431 & \textbf{\textcolor{ForestGreen}{0.8008}} & \textbf{\textcolor{ForestGreen}{100}} & 0.0  \tabularnewline
\bottomrule
\end{tabular}}
\par\end{centering}
\vspace{-15pt}
\end{table}

\subsection{Multiagent and multi-obstacle benchmarking}

We also tested PARM, PARM*, and PRIMER in two different environments: (1) one agent and two obstacles and (2) two agents and two obstacles. To conduct the experiment, we positioned the agents in a \SI{3.0}{\m} radius circle and had them exchange positions diagonally, as shown in Fig. 4. The obstacles are boxes with 0.5-m edges. We set the maximum dynamic limits to \SI{2.0}{\m/\s}, \SI{10.0}{\m/\s^2}, and \SI{30.0}{\m/\s^3} for velocity, acceleration, and jerk, respectively.

We conducted all simulations on an Alienware Aurora r8 desktop running Ubuntu 20.04, which is equipped with an Intel\textsuperscript{\textregistered} Core\textsuperscript{\texttrademark} i9-9900K CPU clocked at $3.60$ GHz with $16$ cores and $62.6$ GiB of RAM.

Table~\ref{tab:sim-benchmarking} compares the average performance of PARM, PARM*, and PRIMER with (1) one agent with two obstacles, and (2) three agents with two obstacles. The metrics used to evaluate the performance are as follows:

\begin{enumerate}
    \item Computation time: the time to replan at each step.
    \item Sucess rate: the rate at the agents successfully reach the goal without collisions.
    \item Travel time: the time it takes for the agent to complete the position exchange.
    \item FOV rate: the percentage of time that the agent keeps obstacles within its FOV when the agent is closer than its camera's depth range.
    \item Number of continuous FOV detection frames: the number of consecutive frames that an obstacle is kept within the FOV of the agent.
    \item Translational dynamic constraints violation rate: the violation rate of the max velocity, acceleration, and jerk.
    \item Yaw dynamic constraints violation rate: the violation rate of the maxi yaw rate.
\end{enumerate}

Note that for PARM and PARM*, we had them generate both 1 and 6 trajectories per replanning and compared their performance. When an agent generates 6 trajectories, it selects the one with the lowest cost. Although 1-trajectory replanning is faster, 6-trajectory replanning is more likely to find a better trajectory. Table~\ref{tab:sim-benchmarking} shows the 1-trajectory replanning approach is significantly faster. However, in more challenging environments with three agents, PARM*'s 1-trajectory approach has a lower success rate compared to the 6-trajectory replanning approach.

Overall, all three methods achieve successful position exchange with similar performance. However, PRIMER outperforms PARM and PARM* significantly in terms of computation time.

In a more complex environment with three agents and two obstacles, both PARM and PRIMER achieve a high success rate, while PARM* falls short of achieving a 100\% success rate. This is because agents spend too much time optimizing their trajectories in PARM*, causing the optimization constraints to become outdated and leading to conflicts during the check and delay check steps. PARM* also suffers from long computation time, resulting in very few generated trajectories passing the check and delay check steps, thus leading to a low success rate.

Fig.~\ref*{fig:sim_simulation_summary} illustrates (a) Computation Time, (b) Travel Time, and (c-d) Trajectory Smoothness. While the 6-trajectory approach finds smoother trajectories, it requires much longer computation time, resulting in increased travel time. The trajectories generated by PRIMER exhibit smoothness comparable to those of PARM*, while PRIMER have significantly faster computation times compared to PARM and PARM*.

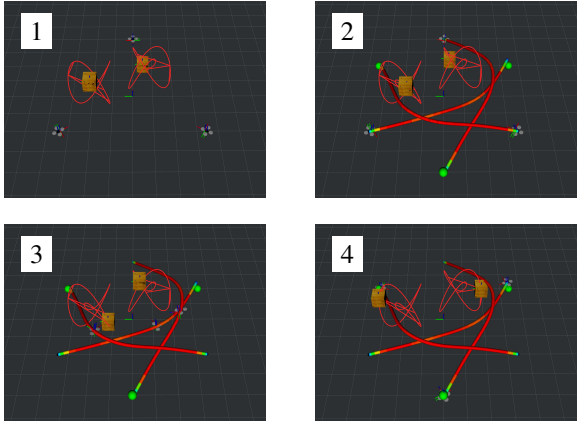
\begin{figure}[h!]
    \centering
    \begin{tabular}{cc}
    \begin{tikzpicture}[every text node part/.style={align=center}]
    \node {\includegraphics[trim={10cm 0 0 10cm}, clip, width=0.4\columnwidth, height=0.3\columnwidth]{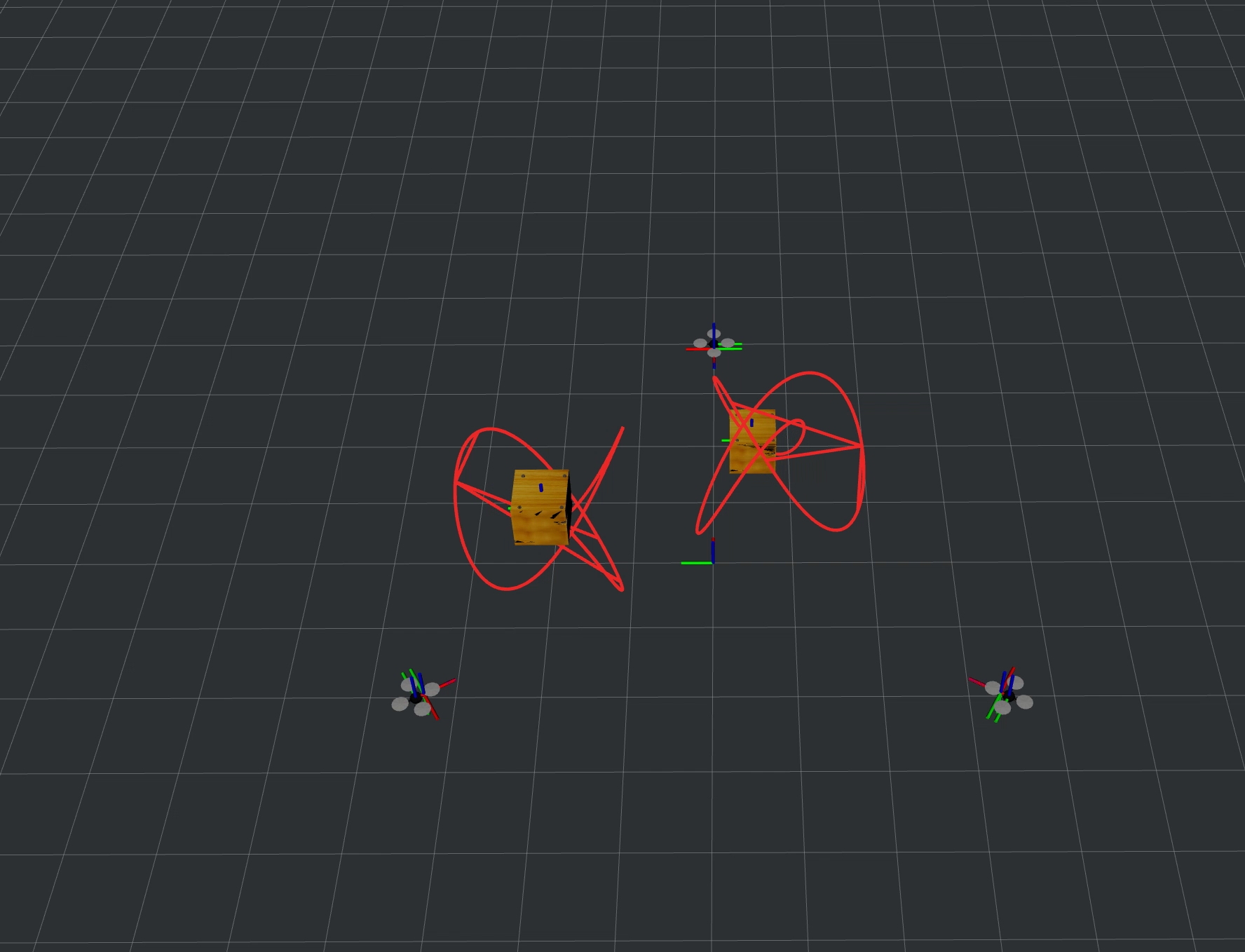}};
    \node [rectangle, fill=white] at (-1.3, 0.9)  {1};
    \end{tikzpicture} &
    \begin{tikzpicture}[every text node part/.style={align=center}]
    \node {\includegraphics[trim={10cm 0 0 10cm}, clip, width=0.4\columnwidth, height=0.3\columnwidth]{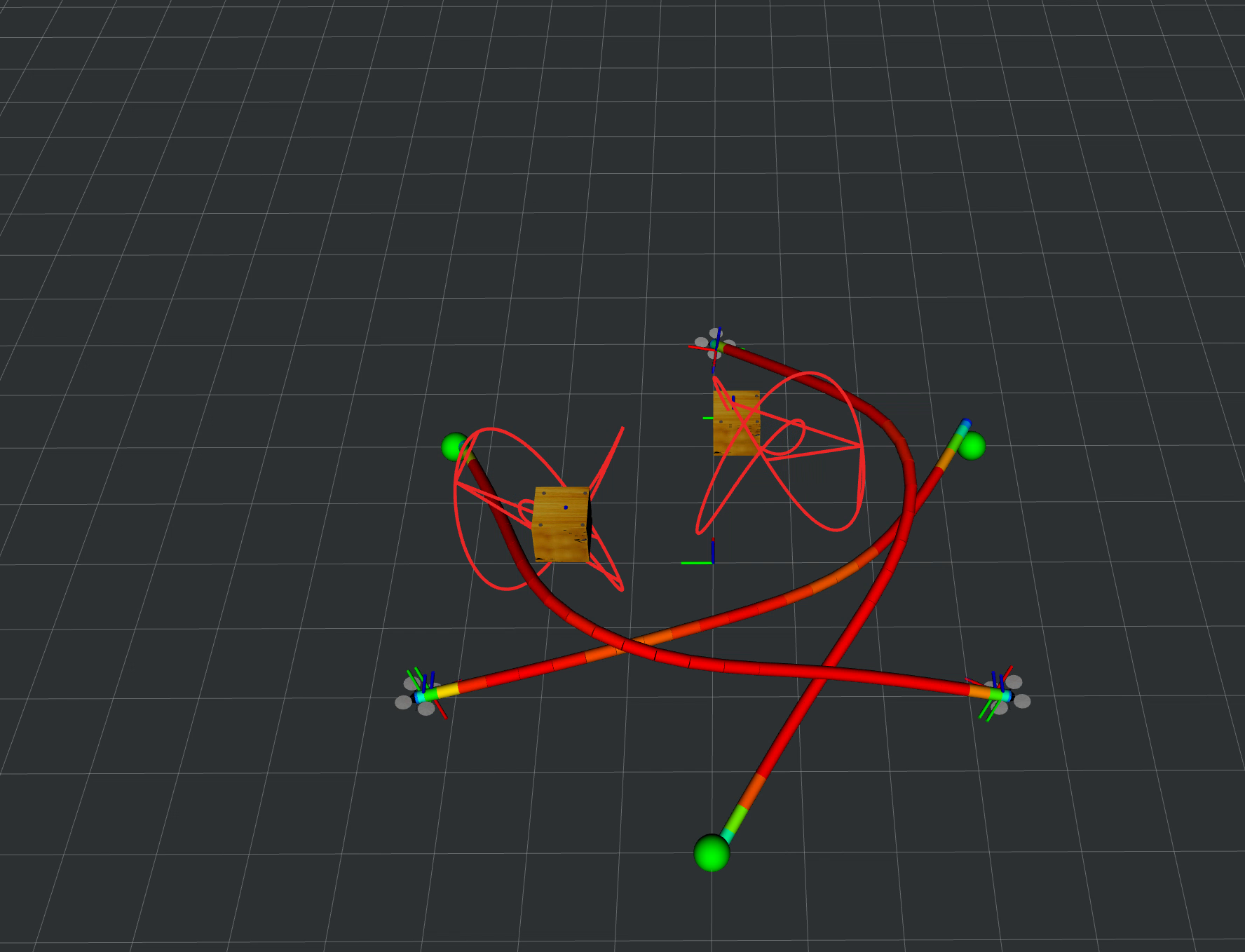}};
    \node [rectangle, fill=white] at (-1.3, 0.9)  {2};
    \end{tikzpicture} \\
    \begin{tikzpicture}[every text node part/.style={align=center}]
    \node {\includegraphics[trim={10cm 0 0 10cm}, clip, width=0.4\columnwidth, height=0.3\columnwidth]{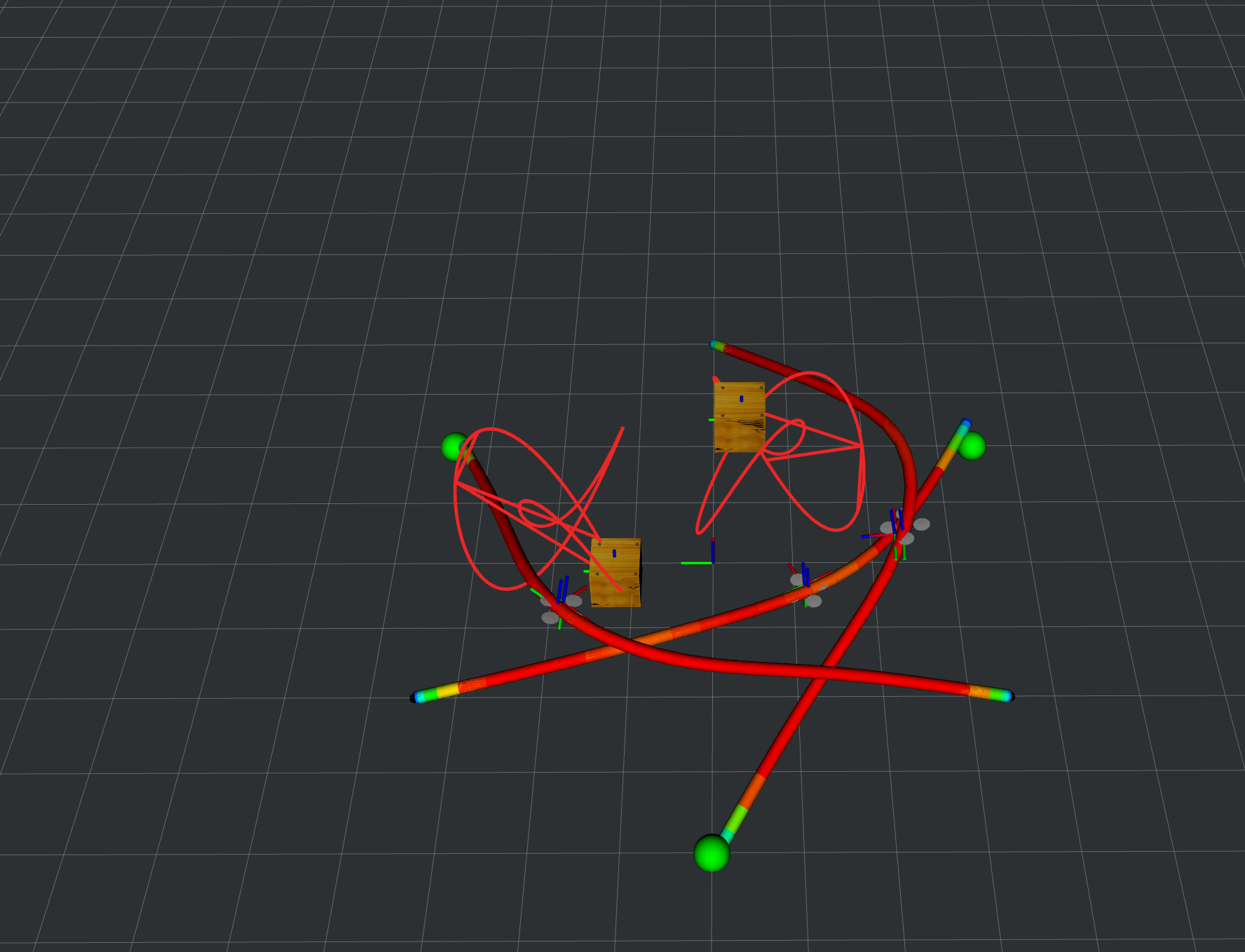}};
    \node [rectangle, fill=white] at (-1.3, 0.9)  {3};
    \end{tikzpicture} &
    \begin{tikzpicture}[every text node part/.style={align=center}]
    \node {\includegraphics[trim={10cm 0 0 10cm}, clip, width=0.4\columnwidth, height=0.3\columnwidth]{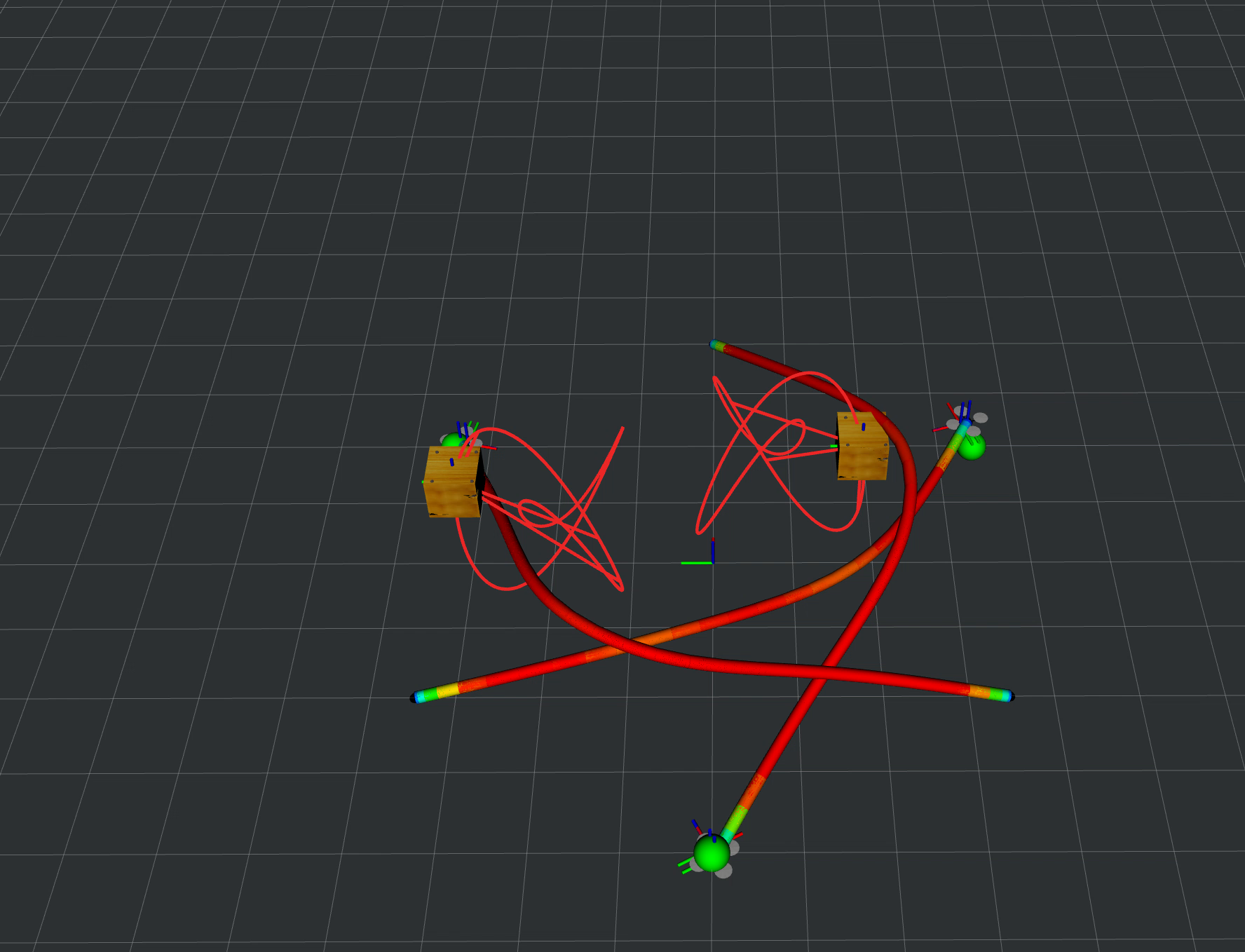}};
    \node [rectangle, fill=white] at (-1.3, 0.9)  {4};
    \end{tikzpicture}
    \end{tabular}
    \caption{Student multi-agent, multi-obstacle, simulation result: We made three imitation learning-based (student) agents fly around two dynamic obstacles. They started at the top-right corner and was commanded to fly to the down-left. For simplicity, we omitted FOV tripods visualization.}
    \vspace{-10pt}
    \label{fig:primer-single}
\end{figure}

\begin{table*}[!t]
    \renewcommand{\arraystretch}{1.4}
    \scriptsize
    \begin{centering}
    \caption{\centering Simulation Benchmarking}
    \label{tab:sim-benchmarking}
    \resizebox{0.8\textwidth}{!}{
    \begin{tabular}{>{\centering\arraybackslash}m{0.1\columnwidth} >{\centering\arraybackslash}m{0.1\columnwidth} >{\centering\arraybackslash}m{0.1\columnwidth} || >{\centering\arraybackslash}m{0.1\columnwidth} >{\centering\arraybackslash}m{0.1\columnwidth} >{\centering\arraybackslash}m{0.1\columnwidth} >{\centering\arraybackslash}m{0.08\columnwidth} >{\centering\arraybackslash}m{0.15\columnwidth} >{\centering\arraybackslash}m{0.15\columnwidth} >{\centering\arraybackslash}m{0.15\columnwidth} }
    \toprule 
    \textbf{Env.} & \textbf{Method} & \textbf{\# Trajs} & \textbf{Avg. Compu. Time [ms]} & \textbf{Success Rate [\%]} & \textbf{Avg. Travel Time [s]} & \textbf{FOV Rate [\%]} & \textbf{Avg. of Max \# Conti. FOV Detection Frames} & \textbf{Translational Dyn. Constraints Violation Rate [\%]} & \textbf{Yaw Dyn. Constraints Violation Rate [\%]} \tabularnewline
    \midrule
    \multirow{6}{*}{\makecell{\textbf{1 agent} \\ \textbf{ + 2 obst.}}} & \multirow{2}{*}{\textbf{PARM}} & 1 & \textbf{\textcolor{Red}{746}} & \textbf{\textcolor{ForestGreen}{100}} & 8.5 & 22.1 & 18.5 & 9.4 & 0 \tabularnewline
    \cline{3-10}
    && 6 & \textbf{\textcolor{Red}{1388}} & \textbf{\textcolor{ForestGreen}{100}} & 7.3 & 25.5 & 15.1 & 10.9 & 0 \tabularnewline
    \cline{2-10}
    & \multirow{2}{*}{\textbf{PARM*}} & 1 & \textbf{\textcolor{Red}{1462}} & \textbf{\textcolor{ForestGreen}{100}} & 7.0 & 26.4 & 31.1 & 0 & 0 \tabularnewline
    \cline{3-10}
    && 6 & \textbf{\textcolor{Red}{3636}} & \textbf{\textcolor{ForestGreen}{100}} & 9.1 & 20.6 & 26.0 & 0 & 0 \tabularnewline
    \cline{2-10}
    & \textbf{PRIMER} & 6 & \textbf{\textcolor{ForestGreen}{140}} & \textbf{\textcolor{ForestGreen}{100}} & 4.2 & 22.0 & 33.0 & 0 & 0  \tabularnewline
    \hline
    \multirow{6}{*}{\makecell{\textbf{3 agents} \\ \textbf{ + 2 obst.}}} & \multirow{2}{*}{\textbf{PARM}} & 1 & \textbf{\textcolor{Red}{728}} & \textbf{\textcolor{ForestGreen}{100}} & 25.7 & 17.1 & 36.0 & 3.2 & 0 \tabularnewline
    \cline{3-10}
    && 6 & \textbf{\textcolor{Red}{1415}} & \textbf{\textcolor{ForestGreen}{100}} & 9.3 & 21.0 & 49.0 & 7.8 & 0 \tabularnewline
    \cline{2-10}
    & \multirow{2}{*}{\textbf{PARM*}} & 1 & \textbf{\textcolor{Red}{1839}} & 60 & 8.5 & 24.0 & 63.3 & 0 & 0 \tabularnewline
    \cline{3-10}
    && 6 & \textbf{\textcolor{Red}{5762}} & 90 & 14.1 & 24.3 & 101.1 & 0 & 0 \tabularnewline
    \cline{2-10}
    &\textbf{PRIMER} & 6 & \textbf{\textcolor{ForestGreen}{109}} & \textbf{\textcolor{ForestGreen}{100}} & 6.6 & 22.6 & 60.2 & 0.3 & 0 \tabularnewline
    \bottomrule
    \end{tabular}}
    \vspace{-5pt}
    \par\end{centering}
\end{table*}

\begin{figure*}
    \centering
    \begin{tabular}{cc}
      \subfloat[\centering Computation Time \label{fig:sim_computation_time}]{\includegraphics[width=0.4\textwidth]{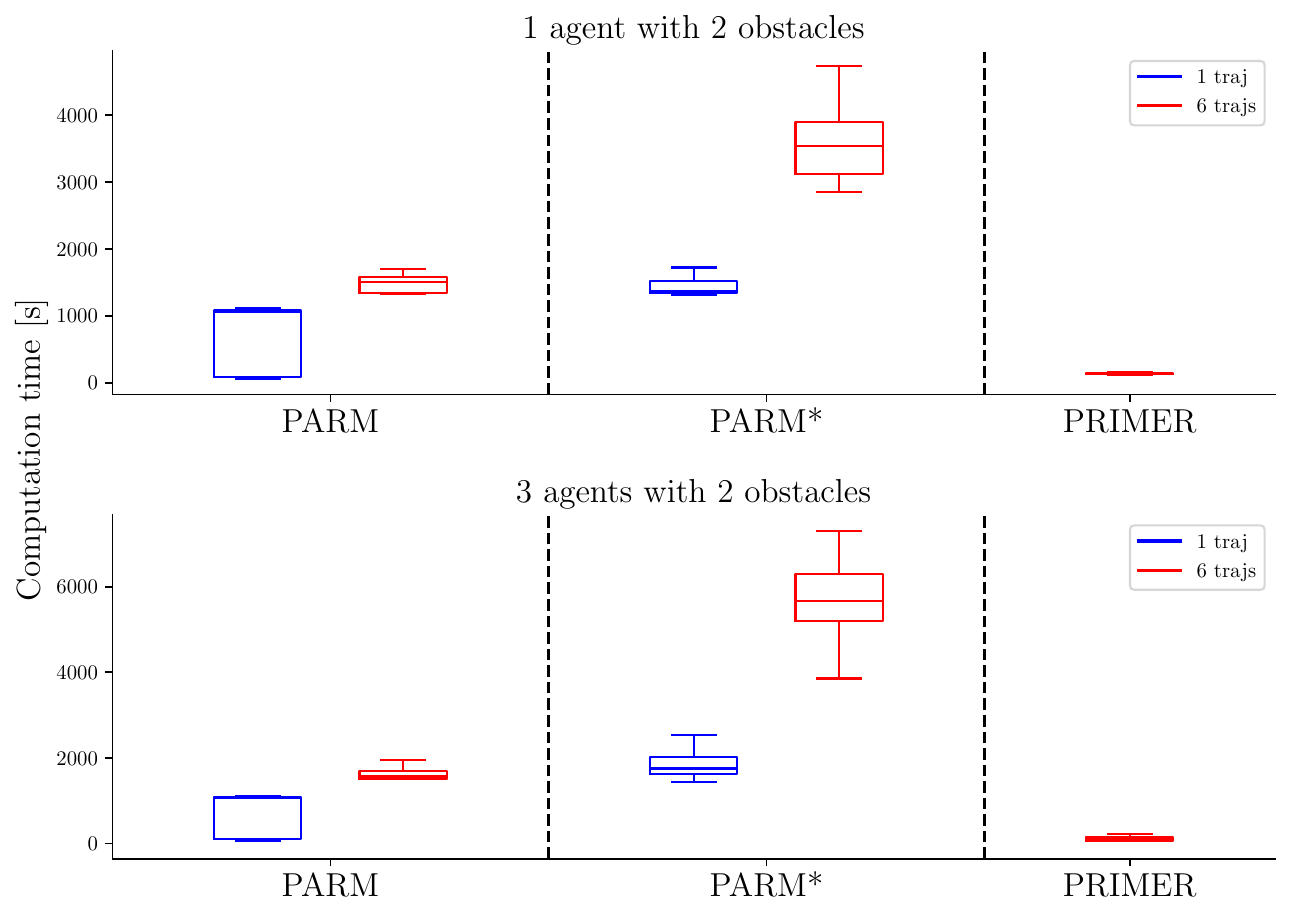}} &
      \subfloat[\centering Travel Time\label{fig:sim_travel_time}]{\includegraphics[width=0.4\textwidth]{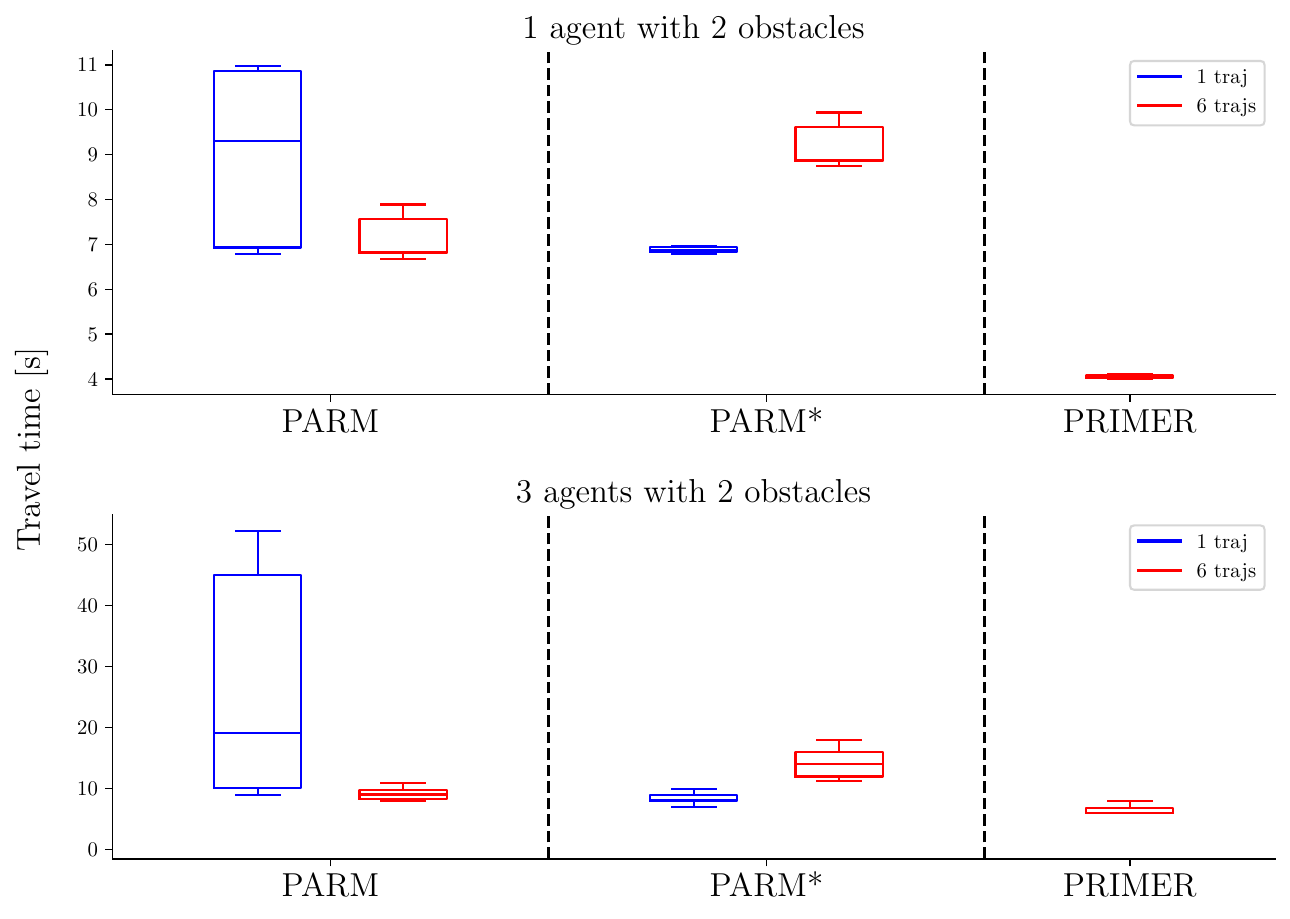}} \\
      \subfloat[\centering Trajectory Smoothness (Acceleration) \label{fig:sim_traj_smooth_acc}]{\includegraphics[width=0.4\textwidth]{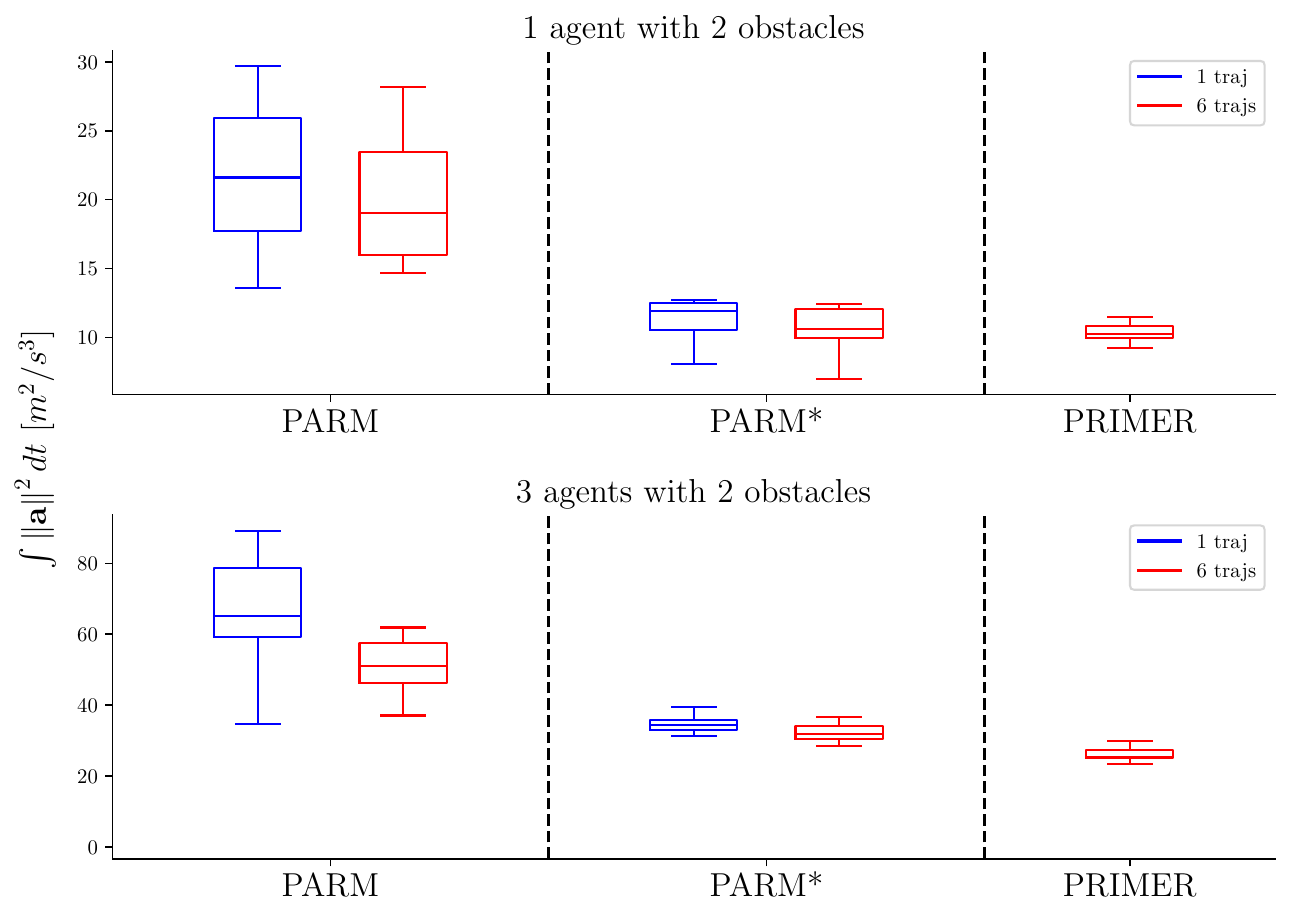}} &
      \subfloat[\centering Trajectory Smoothness (Jerk) \label{fig:sim_traj_smooth_jerk}]{\includegraphics[width=0.4\textwidth]{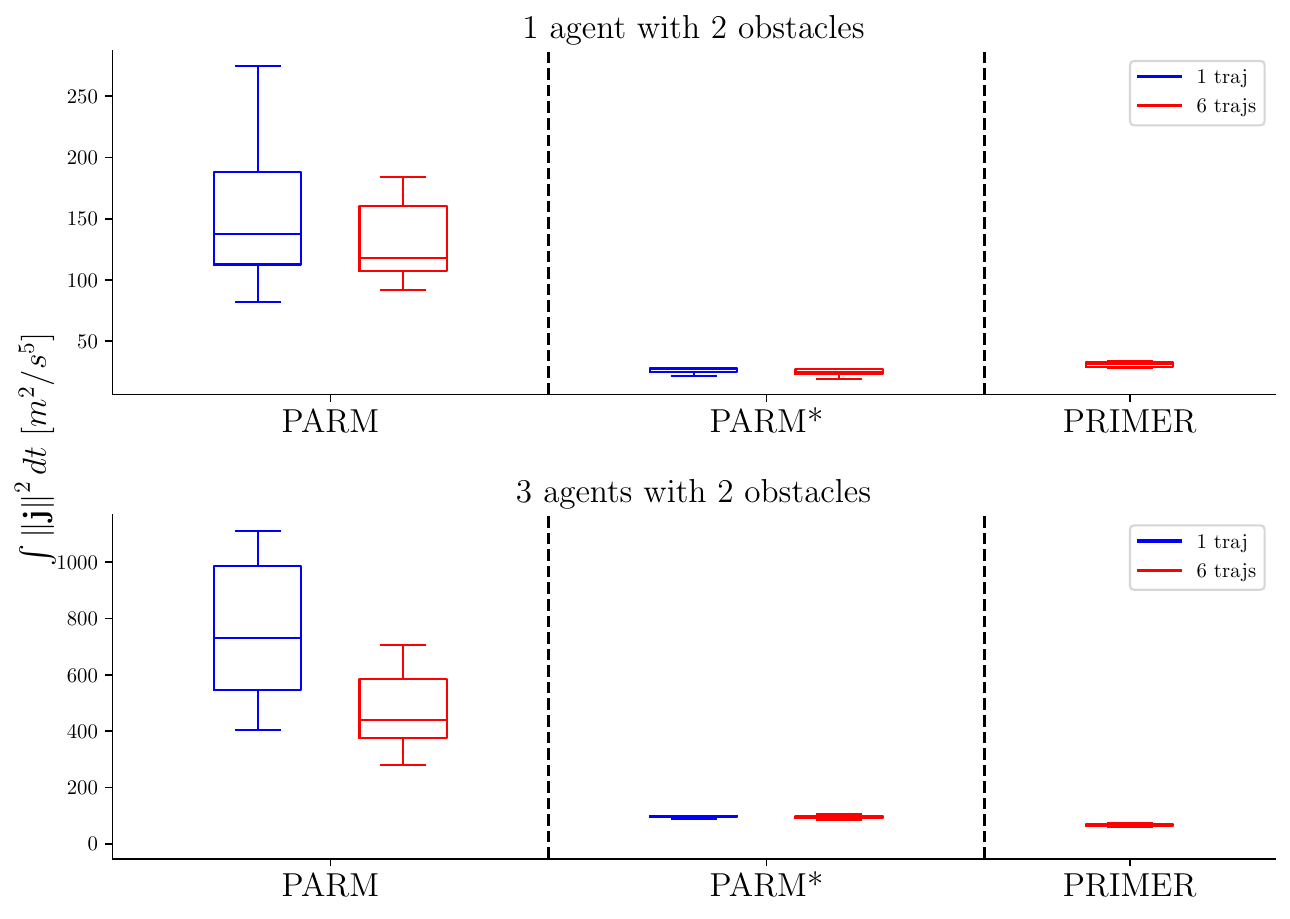}}
    \end{tabular}
    \caption{Results of flight simulations. (a) The student's computation time is much faster than that of the expert, and (b) the student's travel time is also much shorter; this is mainly because of the faster computation time. (c-d) Since the student achieves faster replanning, it does not need to stop as the expert does, and that leads to smoother trajectory generation.} 
    \label{fig:sim_simulation_summary}
    \vspace{-20pt}
\end{figure*}

\section{Conclusions}\label{sec:conclusion}

In conclusion, our work has addressed the critical issue of trajectory deconfliction in perception-aware, decentralized multiagent planning. We first presented PARM and PARM*, which were perception-aware, decentralized, asynchronous multiagent trajectory planners that enabled teams of agents to navigate uncertain environments while avoiding obstacles and deconflicting trajectories using perception information. Although these methods achieved state-of-the-art performance, they suffered from high computational costs, making it difficult for agents to replan at high rates.

To overcome this challenge, we presented PRIMER, a learning-based planner that was trained with imitation learning (IL) using PARM* as the expert demonstrator. PRIMER is a computationally-efficient deep neural network and achieved a computation speedup of up to $5614$ times faster than optimization-based approaches, while maintaining high performance. This speedup enables scalability to a large number of agents, making PRIMER a promising approach for large-scale swarm coordination.

Moving forward, our future work will focus on larger-scale simulations and hardware flight experiments to demonstrate the scalability and performance of PRIMER in complex environments with many agents and obstacles.

\bibliographystyle{IEEEtran} 
\bibliography{main}

\end{document}